# Design of a novel Korean learning application for efficient pronunciation correction

**Francis Jeon**
Information System
Hanyang University
Seoul, 222 Wangsimni-ro, Seongdong-gu, South Korea.

**Minseon Kim**
Educational Technology
Hanyang University
Seoul, 222, Wangsimni-ro, Seongdong-gu, Seoul, Republic of Korea, South Korea.

**Hanseon Joo**
Information System
Hanyang University
Seoul, 222, Wangsimni-ro, Seongdong-gu, Seoul, Republic of Korea, South Korea.

May 4, 2022

## Abstract

The Korean wave, which denotes the global popularity of South Korea's cultural economy, contributes to the increasing demand for the Korean language. However, as there does not exist any application for foreigners to learn Korean, this paper suggested a design of a novel Korean learning application. Speech recognition, speech-to-text, and speech-to-waveform are the three key systems in the proposed system. The Google API and the librosa library will transform the user's voice into a sentence and MFCC. The software will then display the user's phrase and answer, with mispronounced elements highlighted in red, allowing users to more easily recognize the incorrect parts of their pronunciation. Furthermore, the Siamese network might utilize those translated spectrograms to provide a similarity score, which could subsequently be used to offer feedback to the user. Despite the fact that we were unable to collect sufficient foreigner data for this research, it is notable that we presented a novel Korean pronunciation correction method for foreigners.

## 0.1 Introduction

The Korean wave (Hallyu) is the term used to describe the global reach of South Korea's cultural economy, which exports pop culture, entertainment, music, tv dramas, and films [1]. Thanks to the Korean wave, the demand for the Korean language is increasing all over the world. According to the latest statistics, about 160,000 students have enrolled in Korean courses in 2020, with the goal of learning and mastering the language. Furthermore, with about 70 million Korean language speakers, Korean is indeed the 14th most spoken language in the world. In order to help foreign students to learn Korean effectively, the Korean government established the King Sejong Institutes, which have 213 branches in 76 countries [2]. Furthermore, as shown in the graph below, the demand for K-pop and K-drama is continuously increasing, so the demand for Korean is also expected to rise in the upcoming years.

The purpose of this research is to create a novel application for Korean teaching apps specialized in pronunciation correction via voice AI technology. Although learning Korean pronunciation is not challenging since there is no tone and consonant and vowel sounds are simple, double consonant/vowel letters, triphthongs, and L/R pronunciations are obstacles that should be overcome [3]. Therefore, we tend to develop applications that could help people learn Korean pronunciation effectively. Furthermore, as mentioned above, since the demand for learning Korean would keep increasing, this proposed software could aid people all over the world. The rest of this paper will be organized as follows: in the Materials and Methods section, the overall systems of the applications will be discussed. Some of the outcomes of the experiments will be discussed in the results section, and our main findings and summary of this paper will be addressed in the conclusion section.

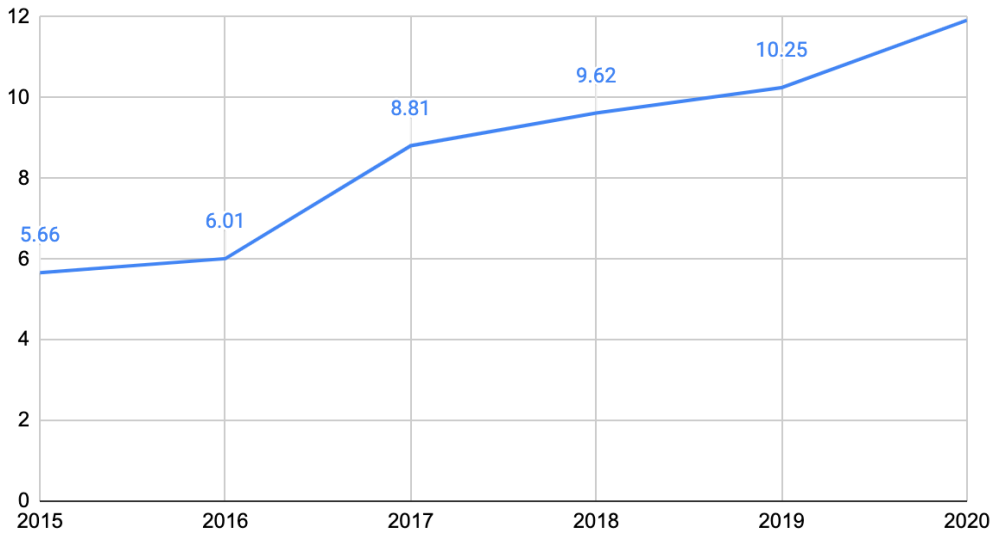

Figure 1: Content Industry Exports from 2015 to 2020

## 0.2 Materials and Methods

### 0.2.1 Proposed System

The overall framework of the suggested pronunciation correction system could be found in the below figure. It consists of three main systems: speech recognition, speech-to-text, and speech-to-MFCC. Speech recognition will be conducted through the user's smartphone, and Google's speech-to-text API would convert the gathered voice into a sentence. Then the generated sentence will be compared against the proper sentence to identify the incorrect component. Furthermore, a librosa, which is a python package for music and audio analysis will be utilized to visualize the waveform of the given data. A similarity score is calculated between the waveforms of the user's voice and proper voice, which could yield the level and top percentage of the user's fluency.

### 0.2.2 Google Speech-to-Text

The Cloud Speech-to-Text API was introduced in April 2017, and the latest version which is based on the novel end-to-end machine learning algorithms supports more than 20 languages and enhances the accuracy of speech recognition. The 'latest short' model is designed for short utterances of a few seconds, which is useful for capturing instructions or other one-shot-directed speech situations [4]. The 'latest long' model is for lengthy material like media or natural speech and dialogues. In terms of fees, speech recognition for audio less than 60 minutes is free. It charges $0.006 every 15 seconds for audio transcriptions that are longer than that [5].

### 0.2.3 Audio Similarity Measure (Siamese Network)

The acoustic similarity will be evaluated by using the Siamese network. This network is a deep learning algorithm that comprises two sub-networks that work together to produce a similarity, and the two sub-networks exchange parameters including weight. The model is made up of a sequence of convolutional layers, each having a single channel, various size filters, and a fixed stride of one. Since the input image is a pair, each image is put into a sub-network to create two outputs, and distance is calculated using the two outputs. To be more specific, the sigmoid activation function is applied to the assessed L1 distance after calculating the L1 distance of outputs from the two sub-networks. Finally, the loss function for the Siamese network is binary cross-entropy. The similarity between the two data is significant if the output value is close to 1, and it is low if it is around 0 [6].

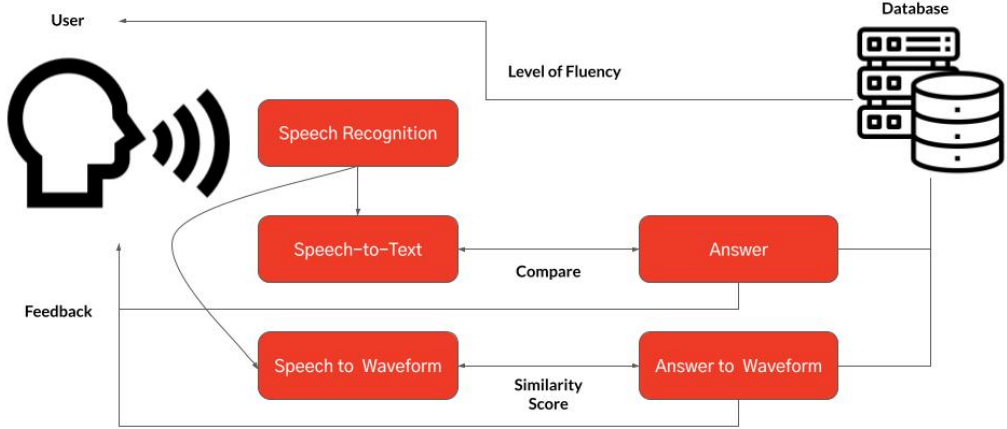

Figure 2: Overall architecture of suggested system

#### 0.2.4  Data Description

Dataset from the AIhub was utilized to yield the expected result of our proposed system, and the dataset is available at `https://aihub.or.kr/opendata/keti-data/recognition-laguage/KETI-02-002` [7]. This dataset was collected using an emotional conversation application, in which users naturally communicated with the application for a certain period of time, and collected data were selected through specific procedures [7].

### 0.3  Expected Result

In order to derive the expected result, simple experiments were conducted. Firstly, the Google sound-to-text was applied to some of the audio files from the given dataset, and this was utilized as the answer in our system. Then we read the same sentence and converted it into a sentence via Google's speech-to-text engine. As a consequence, even though we are native Koreans, it was discovered that if we do not pronounce it correctly, it does not change to the right phrase. Therefore, for the proposed system, the wrong pronunciation could be displayed as a red one, as illustrated in the figure below, and this will make users identify the wrong pronunciations and correct them efficiently.

Answer: 둘 다 청소하기 싫어 귀찮아

User: 요일 날 여기다 청소하기 싫어 귀찮아

Figure 3: Examples of the experimental results of speech-to-text API

As demonstrated below, the same audio files were utilized for the second experiment, and they were successfully converted to Mel-frequency cepstral coefficients (MFCC). The first MFCC representation is a correct description, but the second is inaccurate as mentioned above. With these two spectrograms, the Siamese network could yield the similarity score, which could be further utilized as giving feedback to the user. For instance, if the derived similarity score is higher than 0.9, he or she could be recognized as having the same pronunciation skills as the native Korean speaker. Furthermore, as the range of scores is 0 to 1, the users could easily recognize their ranks among the application users.

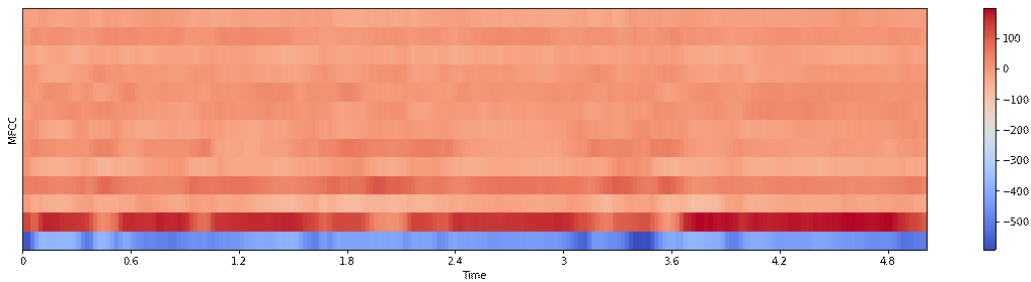

Figure 4: Visualization of converted MFCC from the answer

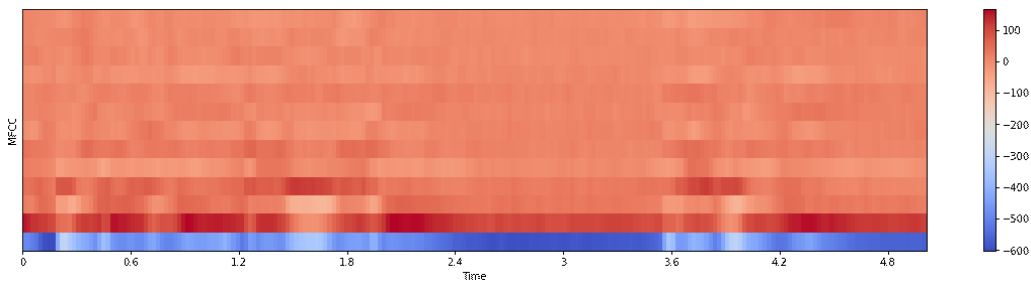

Figure 5: Visualization of converted MFCC from the user

## 0.4 Conclusion

The aim of this research is to propose a novel Korean teaching app, which is suitable for correcting pronunciation more effectively. Speech recognition, speech-to-text, and speech-to-waveform are the three key systems in the proposed system. The user's voice will be converted to both sentence and MFCC through the Google API, and librosa library. Then the app will show the user's sentence and answer, displaying mispronounced parts in red, which will allow users to more effectively identify the wrong parts of their pronunciation. Furthermore, the Siamese network might provide a similarity score using those converted spectrograms, which could then be used to provide feedback to the user. Our paper is significant in that it proposed a new Korean pronunciation correction system for foreigners. However, there still exists a limitation on our paper which should be resolved in further research; since we were unable to collect appropriate data from non-native Korean speakers, it was impossible to effectively implement our suggested application.